\documentclass[conference]{IEEEtran}
\IEEEoverridecommandlockouts
\usepackage{cite}
\usepackage{amsmath,amssymb,amsfonts}
\usepackage{algorithmic}
\usepackage{graphicx}
\usepackage{textcomp}
\usepackage{xcolor}
\usepackage{soul}

\def\BibTeX{{\rm B\kern-.05em{\sc i\kern-.025em b}\kern-.08em
    T\kern-.1667em\lower.7ex\hbox{E}\kern-.125emX}}
\begin{document}

\title{Shaping Multi-Robot Patrol Performance with \\Heterogeneity in Individual Learning Behavior

\thanks{CY is supported by UK Foreign, Commonwealth \& Development Office Services. ZRM is supported by a University of Bristol PhD Scholarship. ERH is supported by the Royal Academy of Engineering under the Research Fellowship program.}
}

\author{\IEEEauthorblockN{Connor York}
\IEEEauthorblockA{\textit{School of Engineering}\\
\textit{Mathematics \& Technology}\\
\textit{University of Bristol}\\
Bristol, UK \\
connor.york@bristol.ac.uk}\\

\and
\IEEEauthorblockN{Zachary R. Madin}
\IEEEauthorblockA{\textit{School of Engineering}\\
\textit{Mathematics \& Technology}\\
\textit{University of Bristol}\\
Bristol, UK \\
zachary.madin@bristol.ac.uk}\\

\and
\IEEEauthorblockN{Paul O'Dowd}
\IEEEauthorblockA{\textit{School of Engineering}\\
\textit{Mathematics \& Technology}\\
\textit{University of Bristol}\\
Bristol, UK \\
paul.odowd@bristol.ac.uk}

\and
\IEEEauthorblockN{Edmund R. Hunt}
\IEEEauthorblockA{\textit{School of Engineering}\\
\textit{Mathematics \& Technology}\\
\textit{University of Bristol}\\
Bristol, UK \\
edmund.hunt@bristol.ac.uk}

}

\maketitle

\begin{abstract}
Individual differences in learning behavior within social groups, whether in humans, other animals, or among robots, can have significant effects on collective task performance. This is because it can affect individuals' response to the environment and their interactions with each other. In recent years there has been rising interest in the question of how individual differences, whether in learning or other traits, affect collective outcomes: studied, for example, in social insect foraging behavior. Multi-robot, `swarm' systems have a heritage of bioinspiration from such examples, and here we consider whether heterogeneity in a learning behavior called latent inhibition (LI) may be useful for a team of patrolling robots tasked with environmental monitoring and anomaly detection. Individuals with high LI can be seen as better at learning to be inattentive to irrelevant or unrewarding stimuli, while low LI individuals might be seen as `distractible' and yet, more positively, more exploratory. We introduce a simple model of the effects of LI as the probability of re-searching a location for a reward (anomalous reading) where it has previously been found to be unrewarding (irrelevant). In simulated patrols, we find that a negatively skewed distribution of mostly high LI robots, and just a single low LI robot, is collectively most effective at monitoring dynamic environments. These results are an example of `functional heterogeneity' in `swarm engineering' and could inform predictions for ecological distributions of learning traits within social groups. 

\end{abstract}


\section{Introduction}

Collective behavior is everywhere in the human and animal world, because cooperation allows social groups to achieve goals that would be difficult or impossible for an individual \cite{Sumpter2010}. For robotics engineers, there are many challenging environments where task complexity or time constraints favor the deployment of cooperative teams of robots \cite{Parker1995}. A distinctive approach to multi-robot systems design is swarm robotics, which takes inspiration from biological self-organization, especially that of social insect societies, to implement decentralized cooperative behaviors \cite{Sahin2005,Schranz2020}. Mathematical and simulation models of large-scale collective behavior have tended to assume homogeneity in agent traits \cite{bonabeau1999swarm,Sumpter2010,Hamann2012}. While this is a logical starting point for scientific investigation and bioinspired engineering, heterogeneity within social groups such as the social insect `superorganism' is increasingly recognized as having important functional benefits \cite{osheawheller2020}. An excellent example of such heterogeneity is observed in honeybee colony (\textit{Apis mellifera}) homeostasis, where genetic diversity favors enhanced nest thermoregulation via variation in temperature response thresholds \cite{Jones2004,Oldroyd2007}. Another likely example of adaptive individual variation is in cognition: honeybees exhibit variation in a learning behavior known as latent inhibition (LI) \cite{Cook2019}. LI is a form of learning to ignore unimportant information -- in the case of honeybees, this could be unrewarding odors \cite{Ferguson2001}. It has been studied in vertebrates with respect to predator recognition \cite{Mitchell2011,Ferrari2008}, for instance, and in humans low LI is associated with attention disorders \cite{Schachar2007}. A recent study of honeybees by Cook et al. exploited the genetic heritability of LI to create artificially manipulated colonies with only high-LI, only low-LI and 50/50 mixed high- and low-LI worker bees. The colonies' collective foraging behavior was then compared to non-selected control colonies \cite{Cook2020}. The study found that high-LI colonies preferred to visit familiar food locations, whereas low-LI colonies visited familiar and unfamiliar locations equally. In mixed-LI colonies, low-LI individuals were influenced by high-LI individuals, via their intensive recruitment efforts, to visit familiar locations. Thus, differences in individual cognition --  attention to important information, i.e. known food locations -- and inter-individual interactions were both found to be significant in driving overall collective behavior, or what might be called `collective cognition' \cite{Couzin2009}. The distribution of LI within natural colonies, and its influence on novelty seeking \cite{Liang2012}, is hypothesized to help manage exploration--exploitation trade-offs over time and space in dynamic environments \cite{Cook2020}. Thus, the variation underlying this distribution is presumably under natural selection \cite{Bhagavan1994}, having important adaptive significance. 

Here, we are inspired by the opportunity for multi-robot systems engineering to introduce variation in individual robot learning behavior, with the strategic aim of obtaining collective learning phenotypes that better manage exploration--exploitation trade-offs. We also consider that such multi-robot systems research could have relevance to understanding whether and how human and animal collectives take advantage of functional heterogeneity \cite{osheawheller2020}. The application domain of our experiments is persistent environmental monitoring for security purposes -- known as robotic patrolling \cite{Basilico2022}. Regular patrolling requires robots to visit important locations as frequently as possible, while adversarial patrolling seeks to introduce an element of unpredictability in movements to make it harder for an attacker to circumvent the system \cite{Huang2019}. Our starting point here is regular patrolling, where one or more robots have to routinely inspect certain sensitive parts of the environment, in return for a `reward' reflecting user preferences on performance. Beyond the typical patrolling problem, we consider that each location has a number of targets that can be selectively inspected for a reward -- thus the task somewhat resembles the well-known exploration--exploitation problem \cite{Kwa2022} in that robots must choose to either inspect known rewarding targets or invest more time in re-inspecting potentially uninformative targets. Locations with clusters of targets could be seen as analogous to co-located `foraging patches' where honeybees might collect food (nectar, pollen). Our main hypothesis is that multi-robot systems that comprise mainly `high-LI' individuals alongside a few `low-LI' individuals, will best embody an effective patrol strategy -- if low-LI individuals can share newly-acquired information. We carry out both agent-based simulations and real robot trials, for various group sizes and individual LI compositions; this is described in section III, Methods, before which we provide more background on LI and robot environmental monitoring in section II. Results are presented in section IV and discussed in section V.

\section{Background}

\subsection{Latent Inhibition and Attention to Relevant Information}

Although latent inhibition (LI) as a phenomenon has been extensively studied over the past 60 years or so, there is not yet a consensus on its mechanism(s), with a variety of theoretical models proposed \cite{Holmes2010}. The typical demonstration of an LI effect involves tracking the acquisition of conditioned response (CR) to a conditioned stimulus (CS) that is followed by an unconditioned stimulus (US). LI is observed as a delay or decrement to the appearance of the CR to a CS that had previously been presented on its own \cite{Holmes2010}. This indicates that something has been learned during initial exposure to the CS that interferes with subsequent learning about the CS, or with the expression of that learning \cite{lubow1959latent}. The most popular explanation of LI is that the pre-exposure to the CS, without any reinforcement, results in a continuing reduction in attention to it; although there could be multiple processes at work to produce LI's effects \cite{Holmes2010}. The mechanisms underlying an individual's cognitive processes of information processing and decision-making are increasingly recognized as having important ecological significance \cite{Dukas1998}, and this is presumably true of biological and robotic collectives as well, especially when one considers the broad scope for realizations of inter-individual variability. The potential of neural network models of adaptive behavior, for example, to be implemented in robots and exhibit natural phenomena such as LI has been noted for some time \cite{Schmajuk1996}, and such effects have been demonstrated in Nao humanoid robots \cite{Billing2014}. For the purpose of our robot model of LI, it seems unnecessary to commit to, and implement, a certain theoretical model of LI -- for our purposes, we are primarily inspired by the potentially useful \textit{effects} of variation in LI. Here the effect of interest, following findings such as Cook et al. \cite{Cook2020}, is variable attention to important (mission-relevant) information; this has also been described in terms of `behavioral persistence' and `site fidelity' in the context of collective foraging \cite{Mosqueiro2017}. In the case of robot patrolling, this could relate to the location of important points of interest, or `hotspots', that ought to be checked regularly. Thus, we implement a simple model of LI, which reflects attention to immediate rewards, as represented by a probability of re-scanning an `irrelevant' (unrewarding) target. This is somewhat akin to high-LI honeybees focusing on familiar food locations that are known to be rewarding. The LI model is described more in the Methods. 

\subsection{Environmental Monitoring and Multi-Robot Patrol}

Robot systems are a popular tool for environmental monitoring across wide spatial and temporal scales, for purposes such as scientific research, pollution tracking, or ecological survey \cite{Dunbabin2012}. Robots are also increasingly used for long-term security monitoring, with the central framing of the problem being multi-robot patrol (MRP), where robots must regularly pass through designated locations of interest \cite{Basilico2022}. The paths between locations are commonly described as a graph, which can be input into a patrolling algorithm that produces target destinations for robots \cite{Portugal2011}. The focus of MRP algorithms is on the effective coordination of robots such that the nodes on the patrol graph are visited as often as possible; and/or on intelligent individual decisions about where to visit next \cite{Portugal2011}. While such algorithms can be deterministic in their outcomes, machine learning methods such as reinforcement learning have also be proposed (e.g. \cite{Santana2004}) which allow for online adaptation to the environment. The focus of the MRP problem on `idleness' minimization (the time between visits to a node) tends to overlook some practical concerns a user might have in a real-world security application. First of all, in an adversarial context, deterministic robot behaviors will be easy to subvert, and so introducing an element of stochasticity to decision-making will be helpful \cite{Huang2019} (one might also wish to benchmark MRP algorithm performance against an attacker model \cite{Ward2023}). In this regard, introducing heterogeneity in robots' behavioral processes could be seen as inherently worthwhile to contribute to system unpredictability. A second practical concern is that locations on the map may require more time investment than a momentary visit. For example, when a robot physically arrives at a location, there may be a choice of how thorough a search to conduct, such as an inspection of one or more entities present at that location, which requires focused attention on that entity for a period of time. For instance, imagine a local search of the corner of an office building: one might have a choice of inspecting a computer monitor, a network switch, a wastepaper basket, or a photocopier. Upon first inspection of the wastepaper basket, there may be nothing `rewarding' present (no anomalous sensor readings), whereas the computer monitor may produce a signal that is distinctive enough to merit a repeat inspection at a later time. This aspect of search thoroughness has not, to our knowledge, been represented in the MRP problem; it could be seen as a particular example of the ubiquitous `exploration--exploitation' problem \cite{Kwa2022,mehlhorn2015unpacking}, although in this case it is not that targets require `exploration' (their presence is known) but rather, re-inspection. This problem can be framed as 'multi-armed bandit' problem (e.g. \cite{Keasar2002}), which when rewards change over time can be referred to as a 'restless bandit' \cite{Speekenbrink2015}. The need to apply selective attention to relevant targets is thus where we perceive the latent inhibition concept to be potentially useful for MRP. When targets are found to be worthy of inspection, one would wish for robots to re-inspect them at a later time, and if they are not, one would wish for robots not to waste time coming back to them frequently at the expense of idleness minimization. However, in a dynamic environment, some level of re-inspection is necessary, to ensure the system's performance does not suffer from stale information \cite{Kwa2022}. A strategy for managing this problem would be to introduce a distribution of selective attention ability, such that most robots preferentially re-inspect rewarding targets. Such group-wide variation could be fixed before deployment, given knowledge about generally effective distributions, or adapt online with robots' experiences. We explain our approach further in the next section.  

\section{Methods}

We carry out experiments with an agent-based patrolling simulator up to $N=6$ robots in an office environment. We describe the implementation of a model of `latent inhibition' in the patrolling robots, with respect to their mission of regularly visiting locations of interest and inspecting appropriate targets. We then describe the simulator and the experimental trials. 

\begin{figure}
    \centering
    \includegraphics[width=0.45\textwidth]{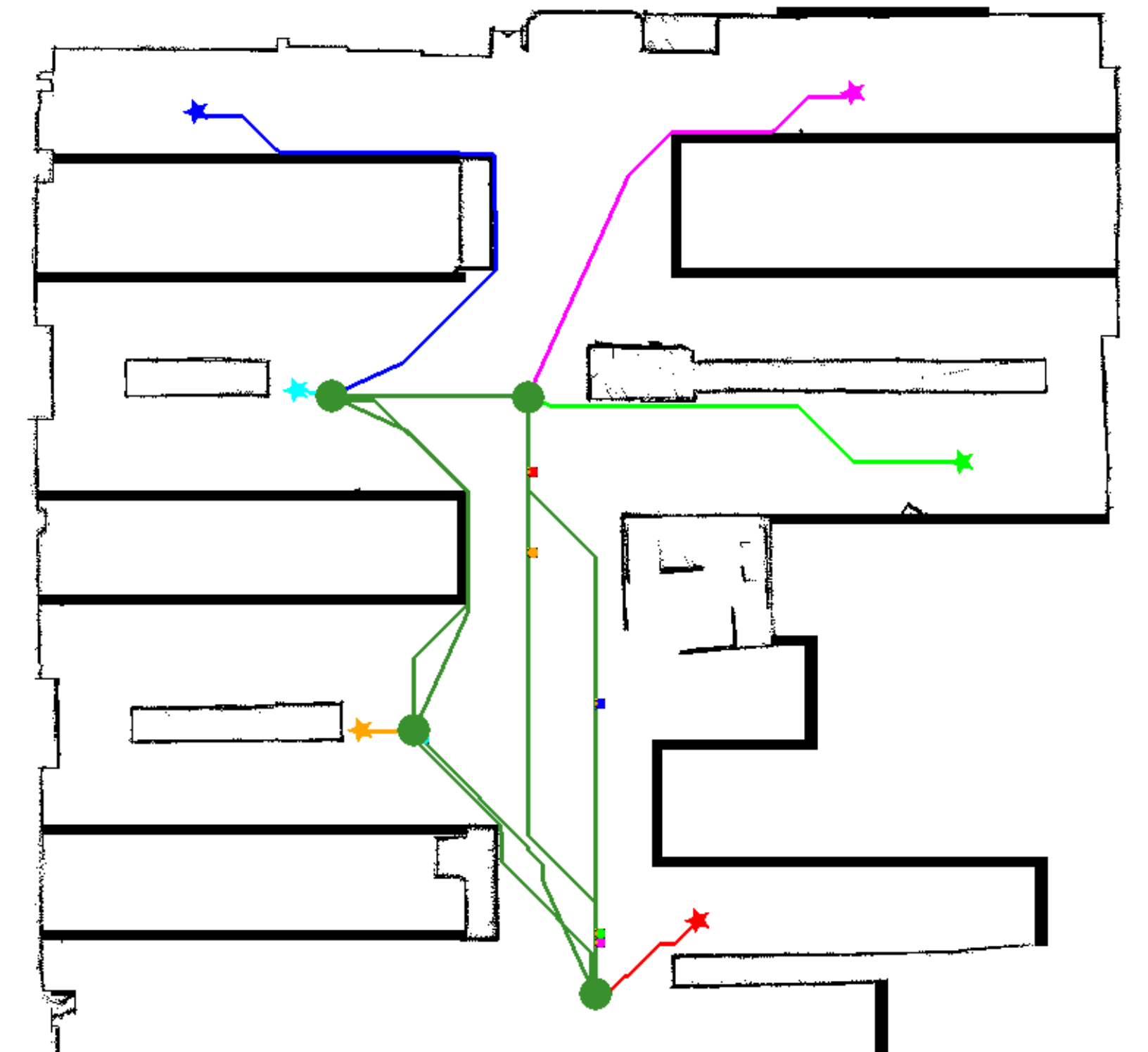}
    \caption{Map and patrol route (dark green) used during the simulations. Stars represent the robots' starting positions, and circles represent the nodes. Colored lines show the robots' paths onto the patrol route for $N=6$. The map was obtained from 2D lidar mapping of our own office environment.}
    \label{fig:patrolmap}
\end{figure}

\vspace{-1pt}

\subsection{Latent inhibition model and patrol strategy}

Robots travel between user-designated points of interest in the environment (waypoints in an ongoing patrol). At each waypoint there are four targets of potential interest that the robot can inspect. The existence of the targets is known but to inspect each one takes an investment of time, representing, for example, a scanning process with the robot's sensors. Here we set a fixed scan time corresponding to around $S_T=8$~s in simulation. Thus, robots have to choose between scanning targets at a certain location or traveling on promptly to the next location.  If a target has never been scanned before by a robot, as is the case for all targets at the beginning of an experimental trial, it will scan it to ascertain whether it is noteworthy (e.g., potential to produce anomalous readings, or vulnerability to attack). To represent this process in the experiments, we set half of the targets to produce a reward ($+1$ reward) when scanned. Robots maintain a memory of whether a target has produced a reward, and if it has produced a reward on the last time of scanning, it will be re-scanned the next time that robot visits the waypoint. If it did not produce a reward, there is a probability $p_r$ of it being re-scanned, which is simply: \begin{equation}
    p_r=1-\text{LI}
\end{equation}
where $\text{LI} \in [0,1]$. That is, low LI individuals have a high probability of re-scanning previously unrewarding targets, and high LI individuals have a low probability of re-scanning them. This represents the narrowly focused attention to immediately relevant information observed in high LI individuals, as seen, for example, in high-LI honeybees' attention to known food sources, or conversely the broader attention to possible rewards of low LI honeybees \cite{Cook2020}. The robots refer to a lookup table of actual rewards, representing the ground truth, which is either fixed for the duration of a trial or may change one or more times, reflecting a dynamic environment. When robots scan a target they immediately and accurately update their belief of whether it is rewarding. 

The robot patrolling algorithm used here is simply a cyclic strategy, i.e. the robot(s) follow a path that visits all nodes on the patrol graph \cite{Portugal2011}. Because we begin with a small patrol graph, all robots follow the same path, unlike a partitioning strategy which would divide the graph up into subgraphs for each robot, which is better suited to maps with separated regions \cite{Portugal2011}. Robots commence a trial physically separated across the map (Fig.~\ref{fig:patrolmap}), but because all of the robots follow the same path, there is scope for robots to catch up and even overtake each other, with consequent scope to interfere with each other's movements. 

\subsection{Agent-based robot simulations}

The Python simulation model is implemented as a two-dimensional grid-world (Fig.\ref{fig:patrolmap}), generated from a real 2D lidar map with each cell representing approximately 2 cm in the real world. On the map locations of interest are represented as nodes within a connected graph. The graph is defined by the adjacency matrix, which is an $W\times W$ matrix where $W$ is the number of nodes in the graph. The weightings between nodes are the lengths of the paths between nodes in the grid-world. In a single time step, robots are able to move in any compass heading from the current cell that they occupy, including diagonal transitions to the eight surrounding cells. Diagonal cell transitions have a $\sqrt{2}$ transition cost, or $1$ otherwise. Each robot occupies a single cell at a time, and executes a patrolling algorithm (here, simply a cyclic patrol route) to determine which node in the graph to visit next, according to the adjacency matrix of the graph. Once an agent has determined which node to visit, a shortest path implementation of A* path planning is used to reach the destination. The simulation was calibrated to produce similar results to an experimental proof-of-concept with real robots (`Leo Rovers') in the same environment: initial trials indicated that 30 minutes of real patrolling corresponded to around 1300 simulation steps. Robots move at 20 cells per step, or around $0.4$ m/s. Physical inter-robot interference is represented by an occupancy rule that only one robot is allowed in a cell at one time. If  a cell that a robot is trying enter is occupied (any cell, not just a waypoint), then the robot will delay for 3 simulation steps, roughly equivalent to 4 seconds, before retrying. As the kinematics of the agents are not modeled, nor the sensor or localization models, the simulator allows fast investigation of multiple system configurations.


We modeled communication between robots by enabling a shared set of beliefs on the reward of targets, which updates immediately when a reward is received. Thus, rather than each robot having a private set of beliefs on the location of rewarding targets, based on its own individual experience, when communication is enabled as soon as a target is found to be rewarding all robots in the group share that belief. This means that a low LI robot can have a significant impact on the group's beliefs, as re-scanning a previously unrewarding target and finding it to be rewarding can not only affect its own future choices about where to scan, but the whole group's choices. 

\subsection{Experimental trials}

\begin{figure*} 
    \centering
    \includegraphics[width=\textwidth]{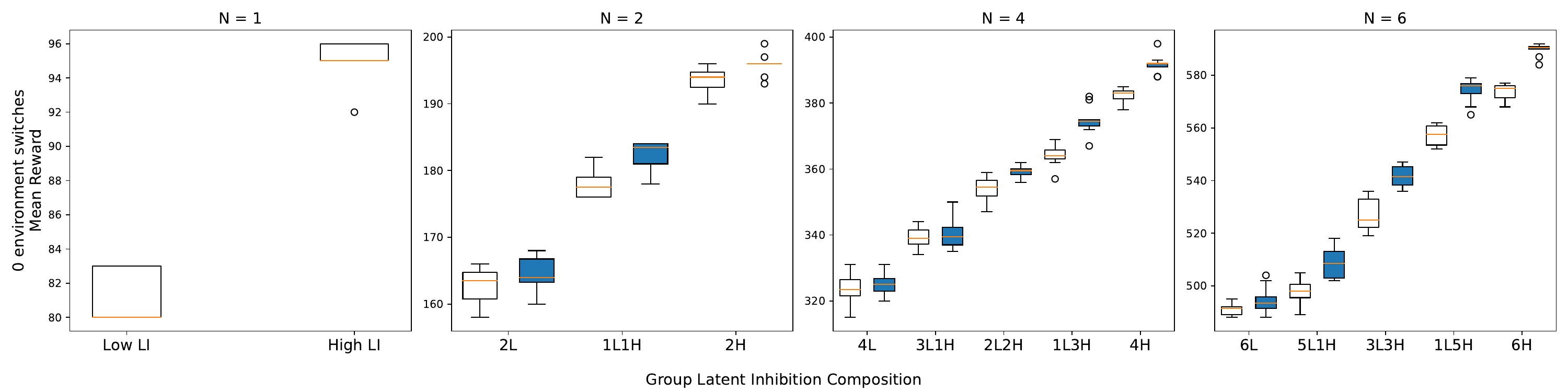} 
    \label{fig:system_1}
    
    \vspace{-23pt} 
    
    \includegraphics[width=\textwidth,trim={0 0 0 0.8cm},clip]{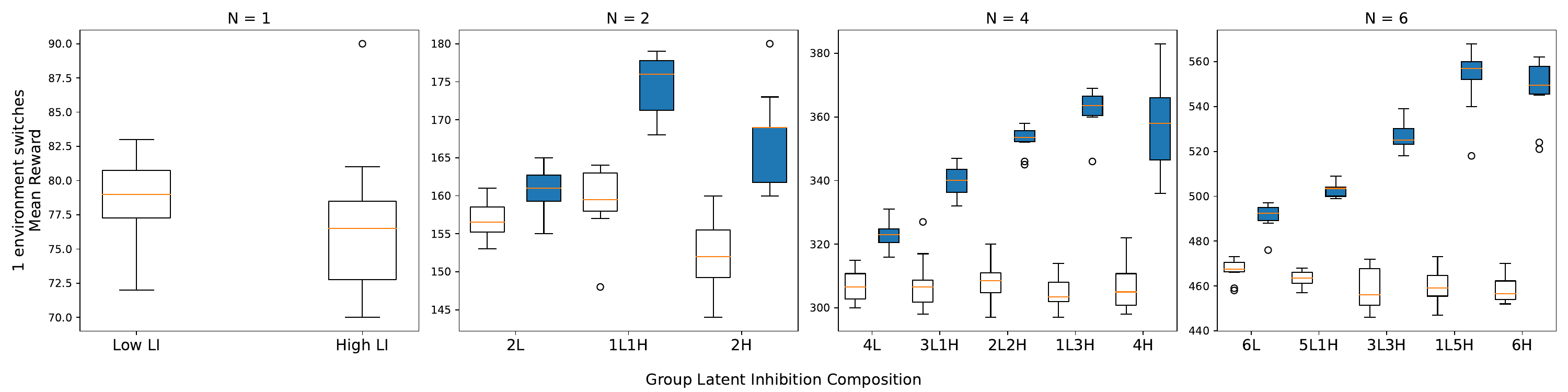} 
    \label{fig:system_2}

    \vspace{-23pt} 
    
    \includegraphics[width=\textwidth,trim={0 0 0 0.8cm},clip]{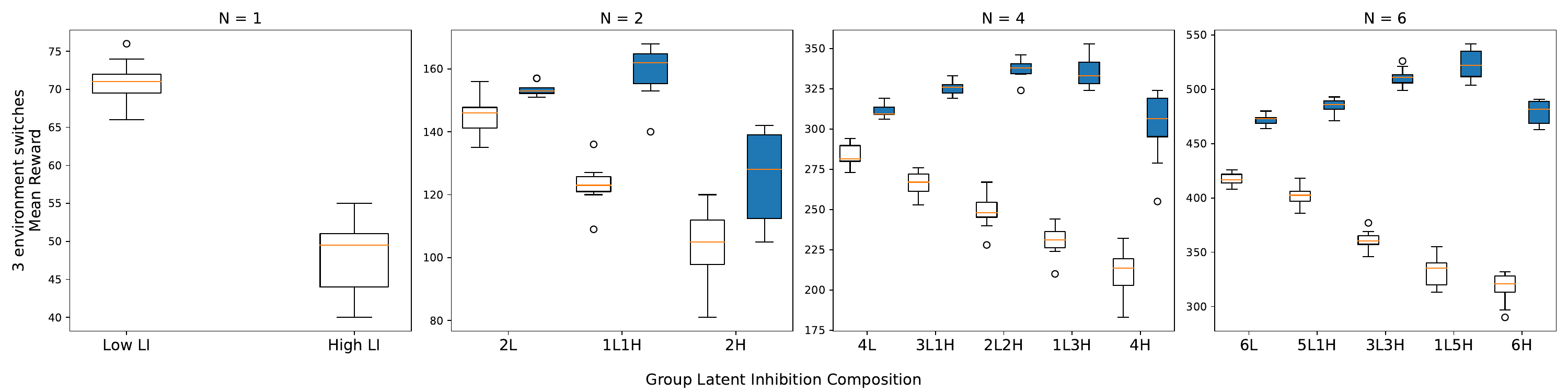} 
    \caption{Average system reward for $N\in\{1,2,4,6\}$ robots by group LI composition. White is communication disabled, blue is communication enabled. Top row: static reward environment. Middle row: 1 environment shift. Bottom row: 3 environment shifts.}
    \label{fig:system_3}
\end{figure*}

The experiment was set up as a cyclic patrol route followed between a certain number of waypoints, $W$, with 4 targets placed at each location. Therefore, there are $4W$ targets available for each robot to scan on each cycle of the patrol route. Of these $4W$ targets, we set half as providing a $+1$ reward, with the other half providing $0$ reward. This assignment is made randomly but is fixed across trials. We used a map of a real-world office environment (a university office space), and set 4 waypoints, so there were a total of 16 targets to be scanned (Fig.~\ref{fig:patrolmap}). To test the effects of a changing environment on the system, in some experimental treatments, after a certain period of time ($T/3$ for 1 switch, or equally spaced $T/(S+1)$ periods for $S$ switches), the rewards assigned to the targets are changed. 50\% switch their reward (from rewarding to unrewarding or vice versa) and 50\% of the tags maintain their previous reward. In simulation trials range from 0 switches (static environment) up to 3 switches (highly dynamic environment). We anticipate that more dynamic environments will favor robot teams comprising one or more low LI individuals, that will be more likely to re-scan previously unrewarding targets. 

For the trials we used $N\in\{1,2,4,6\}$. For variation in our latent inhibition model, we investigated both homogeneous and heterogeneous group compositions of LI, using `low' (L) and `high' (H) LI as $\text{LI}\in\{0.5,0.95\}$. This corresponds to a probability of re-scanning previously unrewarding targets, $p_r$, of $0.5$ and $0.05$ respectively. For heterogeneous compositions we examined either 50/50 mixtures of both types, or a single low or high LI individual among a group of the other type. For example, for $N=4$ we simulated groups of $\{0.5,0.5,0.5,0.95\}$, $\{0.5,0.5,0.95,0.95\}$, and $\{0.5,0.95,0.95,0.95\}$. This was to examine the possible non-linear impact of a single heterogeneous individual on group collective performance. We repeated trials both with and without communication between robots, carrying out 10 simulated trials per configuration, for 1300 simulated time steps. We examine the total reward gathered by the robot system after that time to assess system performance. We use Tukey’s HSD test for pairwise mean performance comparison across system LI compositions. 

\section{Results}

\begin{figure} 
    \centering
    \includegraphics[width=0.5\textwidth,trim={0 0 0 1cm},clip]{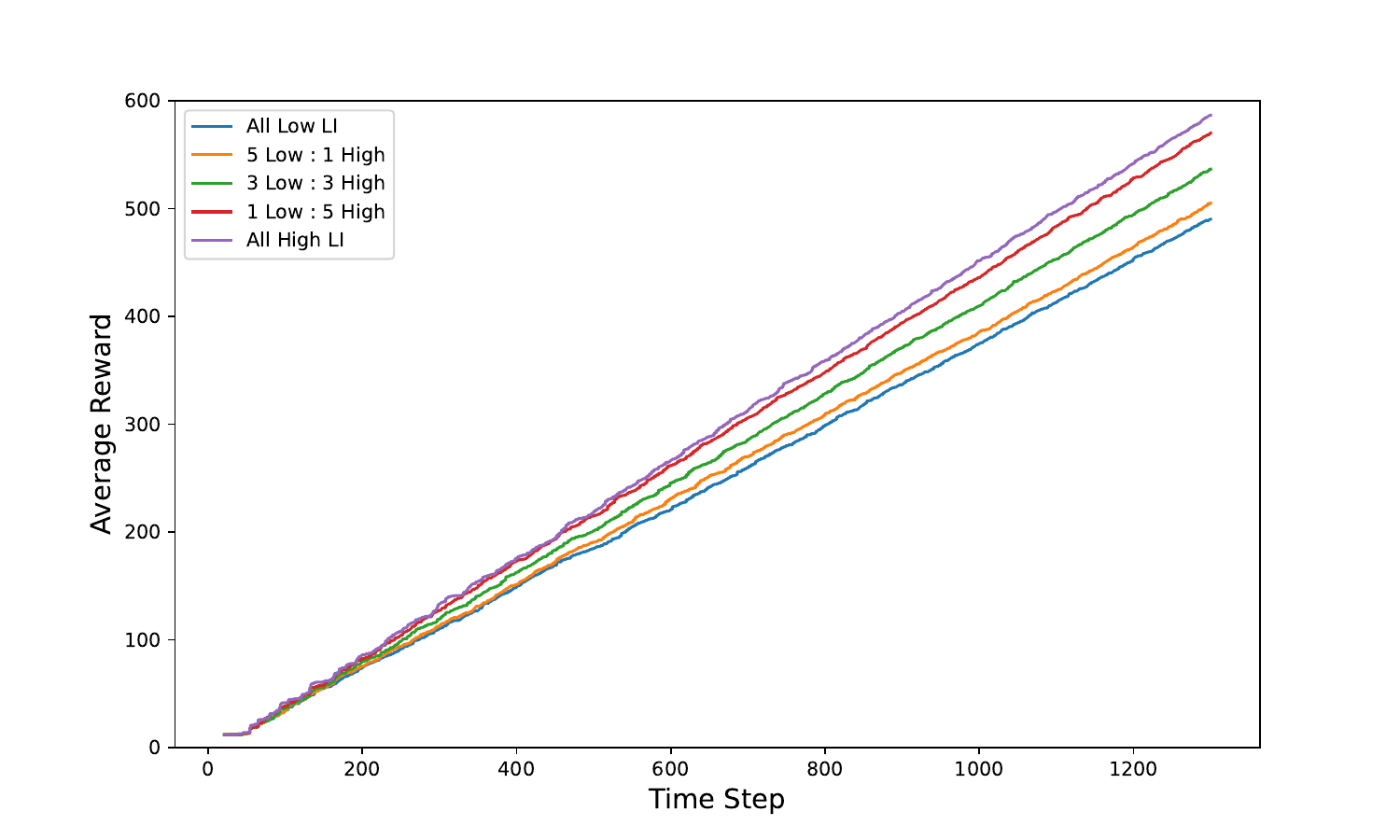} 
    \label{fig:step_1}
    
    \vspace{-23pt} 
    
    \includegraphics[width=0.5\textwidth,trim={0 0 0 1cm},clip]{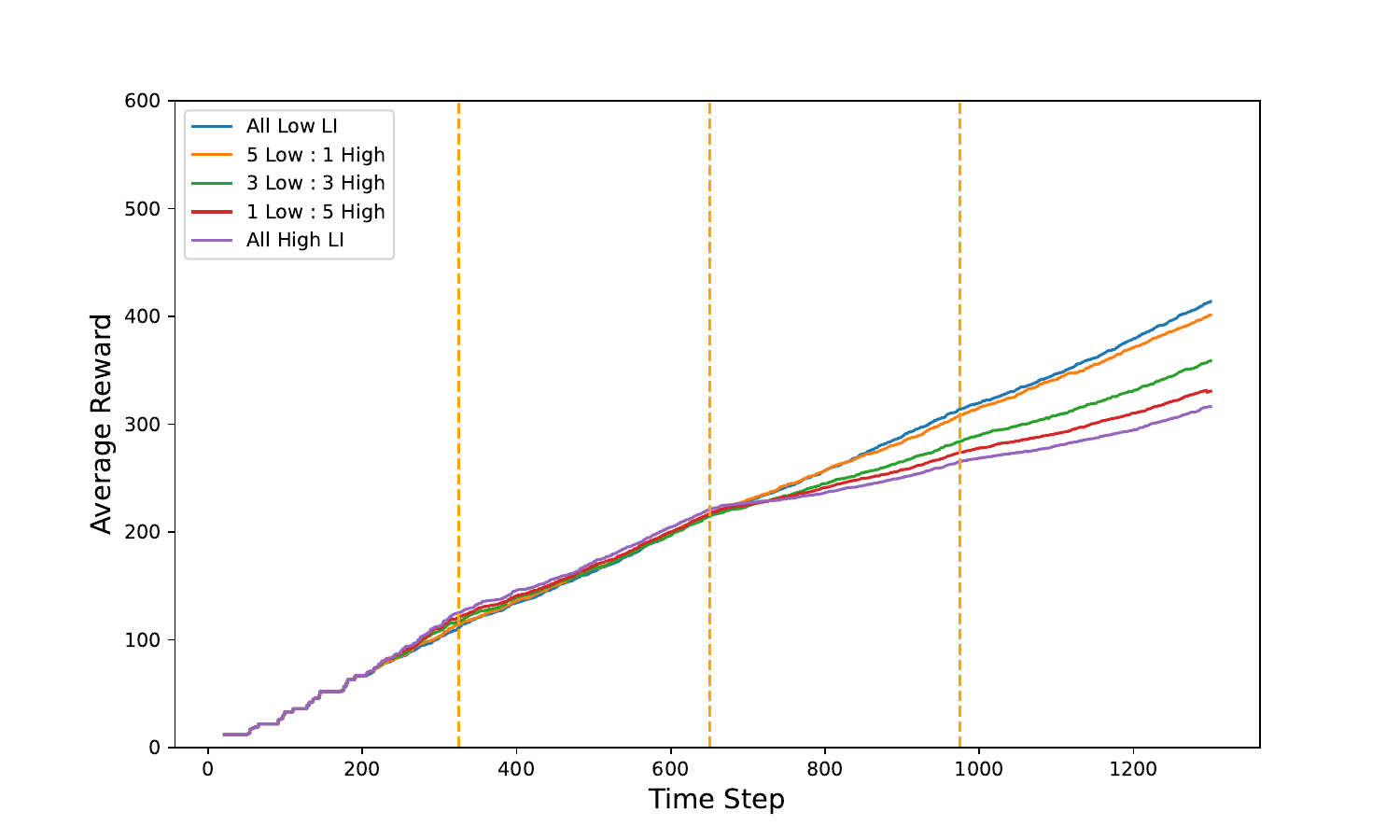} 
    \label{fig:step_2}

    \vspace{-23pt} 
    
    \includegraphics[width=0.5\textwidth,trim={0 0 0 1cm},clip]{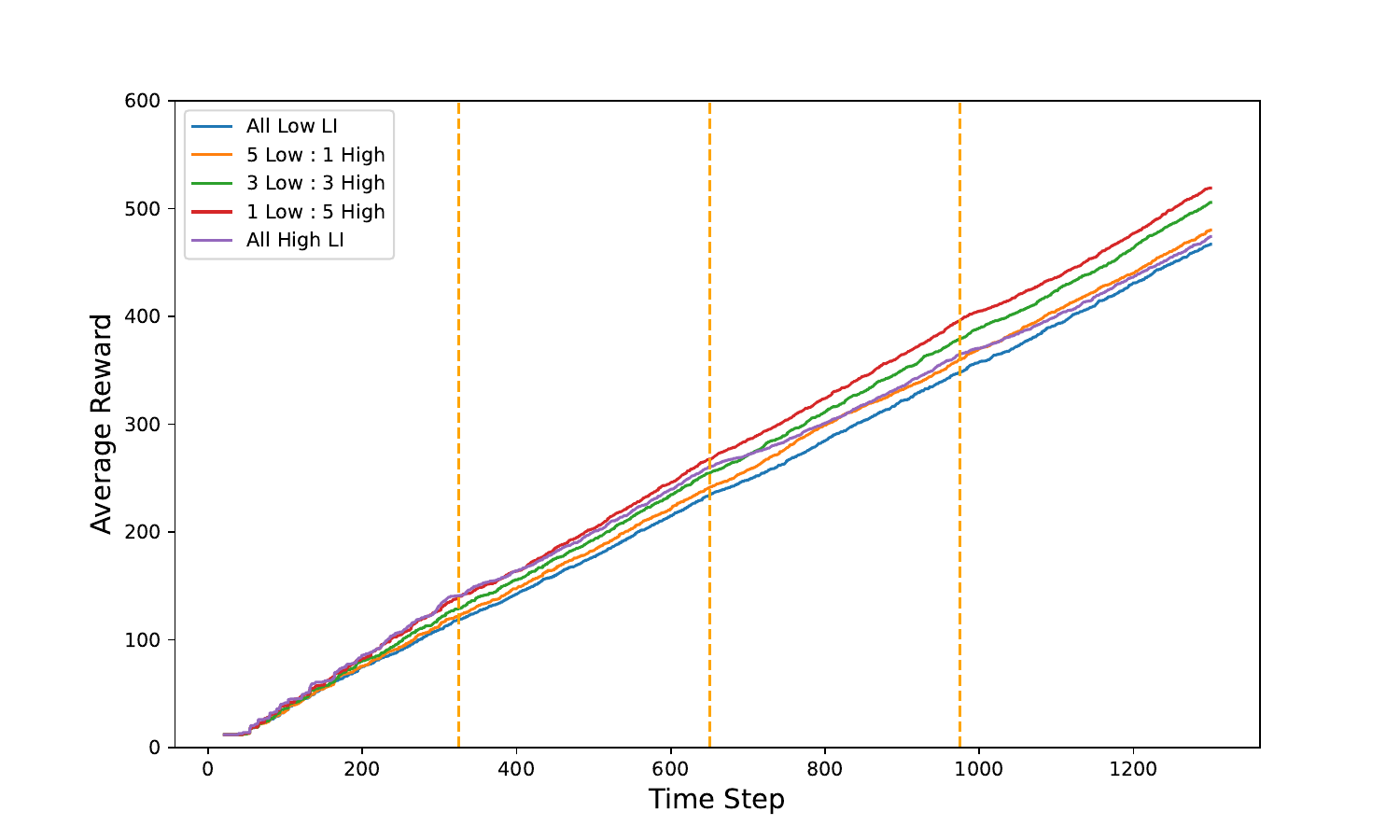} 
    \caption{Average system reward over time for 6 robots. Top: Static reward environment, communication enabled. Middle: 3 environment shifts, communication disabled. Bottom: 3 environment shifts, communication enabled.}
    \label{fig:step_3}
\end{figure}

In a static environment, where the reward status of targets is unchanged for the duration of a trial, individuals or groups with high LI (low attention to previously unrewarding targets) obtain the most reward. This is because there is no benefit to low LI, which only serves to slow robots down in their patrol between waypoints because of frequent, unnecessary re-scanning. There is a small benefit to communication, because early in a patrol robots can share the location of unrewarding targets, thus saving some re-scan time for high LI individuals, allowing them to make faster initial progress around the patrol route (Fig.~\ref{fig:system_3}, top row, Fig.~\ref{fig:step_3}, top). 

When there is a one-off shift in the reward locations (50\% of targets changing: 25\% becoming rewarding and 25\% becoming unrewarding), there can be a benefit to having a lower LI (Fig.~\ref{fig:system_3}, middle row). Starting with the no communication case, for $N=2$ robots, for example, the 2L and 1L1H combination have a higher average total reward than 2H (Tukey post hoc: $Z=5.1, P=0.035; Z=7.6, P=0.001$) though 2L and 1L1H are not different to each other ($Z=-2.5, P=0.41$). Generally though, the advantage of lower LI is not dramatic, because the benefit of successfully identifying newly-rewarding targets is balanced against the slower progress around the patrol route. When communication is enabled, however, the total reward is higher across the board for 1 environment switch, because robots have the benefit of sharing any new information they obtain. Re-examining $N=2$, 1L1H with communication enabled performs significantly better than 2L ($Z=14, P<0.001$), because it has the advantage of one higher LI individual to promptly continue a patrol route between known rewards, including those identified by the lower LI individual; it also outperforms 2H ($Z=7.2, P=0.005$). Examining the larger group sizes, for $N=4$, the compositions with 4H, 1L3H, 2L2H perform better than 4L, 3L1H, although they are not significantly different to each other. For example, 1L3H does not perform better than 2L2H ($Z=9.5, P=0.063$) or 4H ($Z=5.4, P=0.53$). Nevertheless, although 1L3H and 4H have similar means, 4H can be seen as having larger variation in performance (Fig.~\ref{fig:system_3}, $N=4$, middle row). This is because with 4 high LI robots, there is the possibility, though not certainty, for one or more robots to successfully re-scan relevant (changed) targets and share that information with the others. For $N=6$, the group composition with a single low-LI individual (1L5H) is significantly more successful than the 50/50 composition ($Z=26.5, P<0.01$), unlike the $N=4$ team trials. This is presumably because the larger team has the benefit of one low LI individual re-scanning without too many low LI individuals to slow down its overall patrolling between waypoints and known rewarding targets. Even so, for $N=6$, with only a one-off shift in target rewards, it remains the case that 1L5H and 6H are not significantly different in performance ($Z=5.9, P=0.687$) -- the change is not sufficiently disruptive to the robots' knowledge of the environment that six high-LI individuals cannot maintain comparable performance. 

When there is a highly dynamic environment, with three successive shifts in target rewards, the benefit of a negatively skewed distribution of LI is clear for larger group sizes, when communication is enabled (Fig.~\ref{fig:system_3}, bottom row). For $N=6$, 1L5H clearly outperforms 6H ($Z=44.9, P<0.01$) and also the 50/50 composition, 3L3H ($Z=12.4, P=0.047$). Without communication, there is a linear increase in performance the more low LI individuals are present in the group -- because robots are effectively working alone, they need to have low LI to notice the shift in the environment reward distribution (Fig.~\ref{fig:step_3}: compare middle and bottom).

\section{Discussion}

We simulated systems of patrolling robots with communication alternatively enabled or disabled. At each waypoint on the patrol route, there were 4 targets to be scanned, and robots varied in their attention to scanning previously unrewarding targets. This was inspired by the phenomenon of latent inhibition (LI) in humans and animals \cite{Holmes2010,Mitchell2011}, whereby individuals with high LI tend to be better at focusing their attention on relevant information \cite{Cook2020}: in this case, not re-inspecting previously unrewarding targets. Nevertheless, in dynamic environments, sometimes it can be adaptive to reexamine such targets, because their status may have changed with the passage of time. In a social insect system (honeybee colonies), the distribution of LI within natural colonies (and its influence over novelty seeking) is hypothesized by Cook et al. to help manage exploration--exploitation trade-offs over time and space in dynamic environments \cite{Cook2020}. In the context of multi-robot systems or swarm engineering, the impact of variation in such learning behavior could be highly relevant for shaping system strategy in approaching tasks related to such trade-offs. Our main hypothesis was that systems comprising mostly, but not exclusively, high LI individuals, will be generally most effective at monitoring dynamic environments -- if they can communicate freely. This is because the low LI individuals could share information gained from more exploratory behavior, while high LI individuals could focus their attention on known rewarding targets. In natural honeybee colonies, collective behavior is found to resemble that of artificially composed low-LI colonies \cite{Cook2020}, which is coherent with natural selection for a highly dynamic environment. Here, we found that simulated robot systems with compositions of low LI individuals also performed best in highly dynamic environments. When communication between individuals is disabled, there is a clear correspondence between number of low (high) LI individuals and performance in dynamic (static) environments, because individuals are having to rely entirely on private information to make their target choices. In larger group sizes, when communication is enabled, the addition of only one low-LI individual to a group of high-LI robots led to a significant performance increase: an example of the nonlinear impact of heterogeneous individual behavior on collective behavior, an effect which is increasingly recognized as ubiquitous in the natural world \cite{Robson1999,Modlmeier2014a,Brown2014}. 

A recent review of the mechanisms of collective learning by Collet et al. identified three distinct ways through which groups can improve their collective task performance \cite{Collet2023}. These are: individuals learning to better solve tasks alone; members learning about each other's behavior to be better coordinated; and learning to better complement others through specialization of their own behavior \cite{Collet2023}. To obtain the benefits of a suitable LI distribution observed in this study, according to the experienced extent of environmental change, robots could be endowed with some level of plasticity in their LI value. This would be the third form of learning identified by Collet et al. (complementarity via specialization). Because LI is commonly regarded as a stable cognitive trait, adaptability in LI could be regarded as some form of phenotypic plasticity, which we have previously considered in respect of `personality' variation \cite{Hunt2020a,Hunt2020b,Hunt2020c}, another aspect of cognitive variation  \cite{Griffin2015,Sih2012}. Specialization implemented through models such as response threshold reinforcement can be highly effective at improving task performance, yet in dynamic environments flexibility needs to be retained somehow to allow respecialization \cite{Kazakova2020}. In future work we will consider individual flexibility in LI/attention traits for obtaining adaptive group-level distributions and thus achieving effective collective learning in the sort of task we have examined here, such as environmental monitoring. The link between models of individuals' cognitive processes and physical action in collective behavior research is relatively weak, arguably in both biology and robotics \cite{Hamann2012,Zhang2023}, with ample scope to better reflect the diversity and dynamism of individual differences in traits such as learning and personality. We plan to examine this further, including with real robot experiments, which may help to shed light on commonalities in adaptive embodied collective cognition in humans, animals and robots.

\bibliographystyle{abbrv}
\bibliography{latent_inhibition}

\end{document}